\documentclass[conference]{IEEEtran}

\usepackage{amsmath}
\usepackage{array}
\usepackage{booktabs}
\usepackage{cite}
\usepackage{graphicx}
\usepackage{microtype}
\usepackage{url}

\newcolumntype{L}[1]{>{\raggedright\arraybackslash}p{#1}}
\newcolumntype{C}[1]{>{\centering\arraybackslash}p{#1}}

\title{ParEvalLayer: When Partial LLM-Agent Evaluations\\
Support a Decision}

\author{
\IEEEauthorblockN{Wei-Jung Huang}
\IEEEauthorblockA{\textit{Independent Researcher}\\
United States\\
william.wj.huang@gmail.com}
\and
\IEEEauthorblockN{Bonan Shen}
\IEEEauthorblockA{\textit{Independent Researcher}\\
United States\\
shenbonan2@gmail.com}
}

\begin{document}

\maketitle

\begin{abstract}
LLM-agent evaluations often produce task outcomes long before the full benchmark run is complete. A partial score is tempting to report, but it does not show whether the observed tasks support the same conclusion as the completed evaluation. Early tasks can omit important parts of a benchmark, running cheaper tasks first can distort the observed sample, and a rule that decides only easy pairs can appear accurate while leaving many comparisons unresolved. We introduce ParEvalLayer, a decision layer that reads paired outcomes for two agent systems and a comparison policy chosen in advance. For each partial run, it records whether the tested agent system is better by the required amount, is not better by that amount, needs more evidence, or should abstain. We evaluate ParEvalLayer by replaying completed public benchmark data as if each evaluation had stopped earlier. At each point, ParEvalLayer applies the policy using only the outcomes observed so far; if it reaches one of the two comparison judgments, we check whether that judgment matches the completed data for the same system pair. With the main comparison rule, three of the public benchmarks reach the same decision as the completed evaluation after observing only 15\% to 25\% of task outcomes. Other benchmarks require more task outcomes. This variation shows why a partial score alone is not enough: reports should also state the decision rule and how many comparisons remain without a decision.
\end{abstract}


\section{Introduction}

Agent benchmark evaluations often end as a single score, but producing that score can be expensive. A full run may require wall-clock time, API budget, sandbox resources, and human inspection when agent runs fail or need debugging. The practical question often arrives before the full run finishes: from the task outcomes seen so far, can the evaluation report a comparison between two agent systems, or should it keep running tasks?

We use \emph{agent system} to mean the whole evaluated setup that produces task outcomes: the model, prompts, tools, scaffolding, retry logic, and the benchmark verifier. The comparison studied here asks whether one such system scores higher than another by more than a required improvement amount.

A partial score is tempting because it looks like the final report in miniature. It is often not enough to decide whether one agent system is better than another by the amount specified in the evaluation policy. The tasks observed early can miss repositories, difficulty splits, application groups, or domains that would change the completed conclusion. Running cheaper tasks first can reduce cost while showing only a narrow slice of the benchmark. A rule that decides only easy pairs can appear accurate while leaving many comparisons unresolved. What the report needs is therefore not another partial score, but a record that says whether the task outcomes observed so far are enough to support a comparison conclusion.

As a concrete example, suppose a team compares a new system with a previous one on 100 shared tasks and requires a 5 percentage point gain before reporting the new system as better. After 25 observed tasks, the partial score may show a large gain. That number is still hard to interpret if all 25 tasks came from the same repository family, if harder domains have not appeared, or if the rule would abstain on most close comparisons. A decision record makes those conditions explicit: it can decide that the tested agent system is better by the required amount, decide that it is not, request more outcomes, or state that the current budget is not enough.

ParEvalLayer turns that example into a decision layer for an evaluation service. It reads paired task outcomes for the two agent systems, together with a comparison policy chosen in advance. That policy fixes the required improvement amount, task order, coverage requirement, decision-error target, allowed unresolved rate, and task or resource limit. For the current partial run, the layer either decides that the tested agent system is better by the required amount, decides that it is not, requests more task outcomes, or abstains because the budget ended without enough evidence. The corresponding actions are promote, reject, continue, and abstain.

We evaluate ParEvalLayer by replaying completed public benchmark data as if each evaluation had stopped earlier; we call this completed-record replay. For each pair of agent systems, the completed public record defines whether the tested agent system is better by the required amount on the shared tasks. At each budget, we reveal only the task outcomes observed by that point, apply the policy chosen in advance, and compare each promote or reject action with the conclusion from the completed data. Continue and abstain remain unresolved. We use this replay target because released benchmark reports usually expose a fixed task-outcome record rather than a live stopping process.

ParEvalLayer is aimed at in-progress comparisons rather than benchmark compression. Recent task-reduction work asks whether selected subsets can preserve agent rankings across systems \cite{ndzomga2026efficientbenchmarking}. ParEvalLayer asks a different question: for the partial outcomes already visible in a two-system comparison, is there enough evidence to report under a required improvement amount, coverage rule, and task or resource limit?

The benchmark set is chosen to cover different partial-evaluation risks. SWE-bench Lite and Verified test software-repair leaderboards with large shared task sets \cite{jimenez2024swebench,swebenchleaderboard}. AppWorld and OSWorld-Verified test app-use and desktop-control agents with different task-group structures \cite{trivedi2024appworld,xie2024osworld}. tau-bench adds domain labels and cost fields for a cheap-first ordering check \cite{yao2024taubench,barres2025tau2}. Terminal-Bench adds wall-clock timing for terminal workflows \cite{merrill2026terminalbench}. The main task-count simulations use SWE-bench, AppWorld, OSWorld-Verified, and tau-bench because they expose benchmark-provided task groups. Terminal-Bench is used separately to ask whether elapsed-time budgets reduce unresolved comparisons enough to report.

The replay results separate partial runs that are ready to report from cases that remain unresolved. At the 0 pp threshold under the primary policy, ParEvalLayer reaches the same conclusions as completed evaluations after 15\% to 25\% of task outcomes on AppWorld, OSWorld-Verified, and tau-bench. SWE-bench Verified requires 90\% of task outcomes, SWE-bench Lite remains unresolved by 95\% when the policy only asks whether one system scores higher, and the Terminal-Bench wall-clock subset leaves too many comparisons unresolved even at a 95\% elapsed-time budget. The empirical lesson is not that every benchmark can be shortened by the same fraction. It is that partial evaluation needs a decision record: the action must be tied to the required improvement amount, task order, coverage rule, task or resource limit, and unresolved-comparison target.

This paper makes four contributions:
\begin{enumerate}
    \item We introduce ParEvalLayer for partial LLM-agent evaluations. It reads task outcomes from existing evaluation pipelines and records whether the tested agent system is better by the required amount, is not, needs more outcomes, or should abstain.
    \item We evaluate these partial decisions with completed-record replay: replay completed public benchmark data as if runs had stopped earlier, then check whether each judgment that the tested agent system is or is not better matches the conclusion from the completed data. For a given budget grid, the minimum sufficient budget is the smallest tested budget that meets stated decision-error, task-group coverage, and unresolved-comparison targets.
    \item We make partial decisions traceable to the rule that produced them. Each record stores the action, required improvement amount, task-selection policy, coverage status, and bootstrap tail probabilities, so aggregate budget claims can be traced back to the underlying policy.
    \item We show that early reporting is possible in some public records, but not uniformly. At the 0 pp threshold under the paper's primary policy, AppWorld, OSWorld-Verified, and tau-bench reach decisions at 15\% to 25\% task budget. SWE-bench Verified requires 90\%, SWE-bench Lite leaves too many comparisons unresolved by 95\%, and the Terminal-Bench wall-clock check also leaves too many comparisons unresolved at 95\% budget.
\end{enumerate}
\section{Related Work}

Prior work addresses how to reduce evaluation cost, standardize execution, or control risk when stopping or deferring, but it does not specify what a partial benchmark report should contain. Task-subset methods ask whether fewer tasks preserve scores or rankings \cite{ndzomga2026efficientbenchmarking}. Infrastructure such as HAL standardizes how tasks are run and logged \cite{kapoor2025hal}. Sequential and abstention methods study when a procedure can stop, defer, or control risk \cite{wald1945sequential,howard2021confidence,geifman2019selectivenet,mozannar2020defer,angelopoulos2022crc}. ParEvalLayer connects these concerns at the reporting layer: given the task outcomes currently visible for two agent systems, what action should the partial benchmark report contain?

\subsection{Score-Efficient Evaluation}

Score-efficient evaluation asks how many examples, queries, or benchmark tasks are needed to estimate a score reliably. HELM standardizes broad language-model evaluation scenarios and metrics \cite{liang2023helm}. MT-Bench and Chatbot Arena study LLM-as-a-judge and pairwise arena evaluation for chat models \cite{zheng2023judging}. Cer-Eval and FAQ study test-data sufficiency and statistical guarantees for LLM evaluation \cite{wang2025cereval,wu2026faq}. Finite-population sampling tools also apply when benchmark tasks are treated as a fixed population \cite{serfling1974probability,neyman1934representative}.

Within agent evaluation, Efficient Benchmarking of AI Agents studies whether mid-difficulty task subsets can preserve leaderboard rankings under scaffold and temporal shifts \cite{ndzomga2026efficientbenchmarking}. That objective is related but not the same as ours. A rank-preserving subset can make a leaderboard cheaper; it does not tell an evaluation service whether an in-progress two-system comparison can stop, continue, or abstain under a required improvement amount, coverage rule, and unresolved-comparison target.

These methods inform our uncertainty checks, but a partial benchmark report needs more than a score estimate. It must also state the improvement threshold, required task groups, decision-error target, and allowed unresolved rate. ParEvalLayer puts those choices in the decision record, so a reader can distinguish a supported partial decision from a partial score that is missing required evidence.

\subsection{Agent Evaluation Infrastructure}

Agent-evaluation infrastructure runs tasks and records the evidence from which benchmark reports are made. SWE-bench, AppWorld, OSWorld, and tau-bench provide the public task-level records used here \cite{jimenez2024swebench,trivedi2024appworld,xie2024osworld,yao2024taubench,barres2025tau2}. AI Agents That Matter argues that agent evaluation should account for accuracy, cost, downstream usefulness, overfitting, and reproducibility \cite{kapoor2024agentsmatter}. A recent disclosure audit of LLM-agent benchmark papers makes a related point at the paper level: many benchmark reports do not expose enough information about harnesses, inference cost, failure handling, and scoring to support independent interpretation \cite{moghadasi2026disclosure}. ParEvalLayer addresses a different point in the same reporting chain: it records the policy and unresolved status behind a partial comparison.

The Holistic Agent Leaderboard (HAL) is the closest infrastructure comparison \cite{kapoor2025hal}. HAL standardizes evaluation across virtual machines, models, scaffolds, and benchmarks, and reports the cost and implementation pitfalls of large-scale agent evaluation. HAL-like infrastructure can produce task outcomes; ParEvalLayer starts from those outcomes and decides whether the current partial record supports reporting, needs more tasks, or ends in abstention.

Recent harnesses and benchmark suites also broaden what agent evaluation records can contain. General Agent Evaluation, A2Perf, ProSoftArena, TAM Bench, Terminal-Bench, and AndroidWorld add common interfaces, resource measures, reliability-oriented outcomes, professional software tasks, task coverage, or software-environment traces \cite{bandel2026generalagent,uchendu2025a2perf,ai2026prosoftarena,jia2025tambench,merrill2026terminalbench,rawles2024androidworld}. These systems improve what an evaluation pipeline can log; ParEvalLayer asks when a partial record produced from such logs is enough to report a comparison. We use Terminal-Bench separately for a wall-clock resource check because the primary task-count analyses require benchmark-provided task groups.

Trace-level cost systems such as ClawTrace are complementary because they explain trajectory cost rather than deciding whether a partial comparison is ready to report \cite{yuan2026clawtrace}.

\subsection{Stopping, Abstention, and Risk}

Sequential testing, confidence sequences, abstention methods, and conformal risk control provide tools for stopping under uncertainty, deferring predictions, or controlling risk when predictions are withheld \cite{wald1945sequential,howard2021confidence,geifman2019selectivenet,mozannar2020defer,angelopoulos2022crc}. They do not by themselves specify what a partial benchmark report should contain. ParEvalLayer uses these ideas to answer a reporting question: which action should be reported for a partial two-system comparison, and which rule produced it?

ParEvalLayer uses abstention as a benchmark-reporting action rather than a prediction outcome. For each completed public record, the simulation hides later task outcomes and asks whether the outcomes observed so far lead to the same conclusion as the completed evaluation under thresholds, coverage rules, and abstention targets chosen before those later outcomes are revealed. If not, the decision record states continue or abstain instead of forcing promote or reject.
\section{ParEvalLayer}

\begin{figure*}[t]
    \centering
    \includegraphics[width=0.98\textwidth]{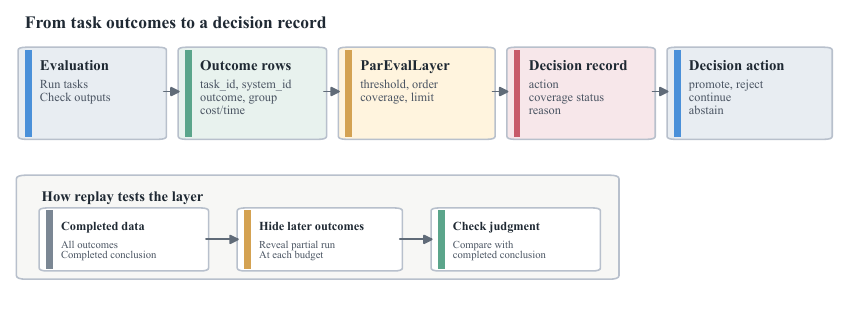}
    \caption{ParEvalLayer reads task outcome rows and a comparison policy chosen in advance, then writes a decision record for the current budget.}
    \label{fig:decision-layer}
\end{figure*}

ParEvalLayer answers a narrow reporting question: do the task outcomes observed so far justify a comparison decision now? Figure~\ref{fig:decision-layer} shows the contract. An evaluation service writes task-level outcomes; ParEvalLayer reads those outcomes, applies the comparison policy chosen in advance, and writes a decision record for the current budget.

The contract has four parts: input and output fields, the completed-record decision used in replay, the actions and aggregate error rates, and the coverage rule and bootstrap check used by the primary policy.

The decision record has four actions. Promote reports that the tested agent system is better than the baseline by more than the required amount; reject reports that it is not better by that amount; continue requests more task outcomes; and abstain states that the budget ended without enough evidence for either final action. In simulations from completed public records, promote and reject are checked against the completed-record decision, while continue and abstain remain unresolved.

\subsection{System Interface}
\label{sec:system-interface}

ParEvalLayer keeps the input contract small: one row per task outcome, with the system identifier, task identifier, outcome, optional task group, optional resource fields, and provenance. Once both systems have an outcome for the same task, the layer forms the paired difference and applies the decision policy chosen before the run reaches the current task or resource limit. The decision record reports the action, number of observed tasks, coverage status, margin, and uncertainty values used to produce that action.

This contract keeps task execution and stopping reports separate. HAL or a similar execution system can manage virtual machines, logs, retries, raw-trajectory scoring, and execution cost; ParEvalLayer starts from the task outcomes those systems write. Using the same task-outcome representation for individual decisions and aggregate summaries keeps each budget claim tied to the decision records that produced it.

At each simulated partial run, the layer reads visible task outcomes up to a task or resource limit and writes a decision record with the action, coverage status, partial margin, and uncertainty check. Completed-record fields are used only for after-the-fact replay checks, not as inputs to the stopping rule.

\subsection{Decision Target}
\label{sec:decision-target}

Each comparison is paired at the task level. Whenever both outcomes are available, the same task compares the two systems. Public leaderboard records do not always provide exactly the same task set for every entry, so the comparison is defined on the shared usable overlap for each ordered pair.

For an ordered comparison, \(A\) is the tested agent system and \(B\) is the comparison agent system. Let \(T_{A,B}=\{t_1,\ldots,t_{N_{A,B}}\}\) be the tasks with usable outcomes for both systems after source-specific filtering. Tables report median pair overlap when \(N_{A,B}\) varies across pairs. For readability, we write \(T\) and \(N\) below when the ordered pair is clear.

Each task outcome is a normalized score, \(Y_{i,A},Y_{i,B}\in[0,1]\). Binary outcomes are the common case. When a public record contains repeated runs for the same system and task, we average those runs before forming pairs. The paired task difference is
\begin{equation}
    D_i = Y_{i,A} - Y_{i,B},
\end{equation}
and the paired improvement on the completed shared task record is
\begin{equation}
    \Delta_{\mathrm{full}} = \frac{1}{N}\sum_{i=1}^{N} D_i.
\end{equation}
We use \emph{budget} for the task or resource limit at which a partial run is evaluated. In the task-count analyses, \(b\) is a fraction of the completed shared task set; in the resource check, the same notation refers to a fraction of elapsed time. For a subset \(S_b\) observed by budget \(b\), the partial paired improvement is
\begin{equation}
    \widehat{\Delta}_{b} = \frac{1}{|S_b|}\sum_{i:t_i\in S_b} D_i.
\end{equation}

Before a partial comparison, the improvement threshold \(\delta\) is chosen in advance. On the available overlap, the completed-record decision is promote when \(\Delta_{\mathrm{full}}>\delta\) and reject otherwise. A 0 percentage point (0 pp) threshold asks only whether \(A\)'s paired success rate is strictly higher than \(B\)'s. Positive thresholds, such as 5 pp or 10 pp, require a larger margin before promotion. The completed-record decision is defined on the shared public overlap for the ordered pair; tasks missing from either entry are not part of that comparison.

This replay target matches what many benchmark reports release: a fixed task-outcome record with a stated comparison rule.

The ordered comparison matters. \(A\) versus \(B\) asks whether the tested system is better than the comparison system by more than the threshold. \(B\) versus \(A\) asks the reverse question. The main analysis therefore keeps ordered comparisons, and we separately test what changes when only one direction is kept from each unordered pair.

\subsection{Actions and Decision Records}
\label{sec:actions-records}

At a given budget, the policy receives the observed subset \(S_b\). Before the configured budget limit, it can continue evaluation. At the limit, it can promote \(A\), return reject, or abstain. A promote action means the observed tasks support the conclusion that the partial paired difference exceeds \(\delta\). A reject action means the observed tasks support the conclusion that the partial paired difference does not exceed \(\delta\). Continue and abstain state which task groups or evidence conditions are still missing.

We measure false promotions and false rejections against the completed-record decision. A false promotion occurs when the partial policy promotes \(A\) but the completed-record decision is reject; a false rejection occurs when the partial policy returns the reject action but the completed-record decision is promote. Abstentions remain in the denominator for the class assigned by the completed record, but they count as neither correct decisions nor decision errors. Otherwise a rule could decide only easy cases and appear accurate while leaving most comparisons unanswered.

For aggregate rates, one evaluation case is an ordered pair \(p=(A,B)\), a task order \(r\), and a budget \(b\). We compute false-promotion rates over cases where the completed-record decision is reject, and false-rejection rates over cases where it is promote. Coverage failure and abstention rates are averaged over all cases.

We read directed pairs as decision instances, not independent systems. The two directions of a pair share outcomes, but they answer different one-sided questions.

Aggregate summaries are derived from per-comparison decision records. A reader can inspect the decision threshold, task policy, missing task groups, and abstentions behind each partial action rather than only a plotted budget curve.

\subsection{Task-Group Coverage}
\label{sec:coverage}

Task-group coverage prevents one specific mistake. A partial run can appear decisive inside a narrow slice while missing groups that would change the decision from the completed record. The groups come from public metadata: repository for SWE-bench, split by difficulty for AppWorld, application group for OSWorld-Verified, and domain for tau-bench. In the equations, we call these groups strata.

The coverage rule asks a simple question before any final action: has the partial run seen enough tasks from every required group for this budget?

For each ordered pair, coverage is computed on the shared task set \(T_{A,B}\). Let \(q_i\) be the stratum for task \(t_i\), \(n_g\) the number of shared tasks in stratum \(g\), \(b\in(0,1]\) the budget fraction, and \(k_b=\lceil bN\rceil\). Strata absent from a pair's shared overlap are excluded for that pair. The primary required count for stratum \(g\) is
\begin{equation}
    r_g(b)=\max\{1,\min\{n_g,\lfloor k_b n_g/N\rfloor\}\}.
\end{equation}
An observed subset satisfies task-group coverage when it contains at least \(r_g(b)\) tasks from every required group. The rule has two roles: every nonempty group must appear at least once when possible, and larger groups receive proportionally larger required counts. For example, if \(N=100\), \(b=0.20\), and three groups contain 50, 30, and 20 shared tasks, the required counts are 10, 6, and 4. If \(\sum_g r_g(b)>k_b\), the target is impossible at that budget and the rule cannot issue a final decision.

A coverage failure occurs when a policy returns promote or reject before satisfying the required counts. The rule deliberately favors a simple reporting check over an optimal task-allocation procedure: it makes missing-slice failures visible before a partial comparison is reported.

\subsection{Decision Rule and Sufficiency}
\label{sec:decision-rule}

The primary policy applies the coverage rule before it considers the score margin. The bootstrap rule checks uncertainty over the observed paired task differences. We use it because the layer sees paired task outcomes and does not require a parametric model for their observed differences. Section~\ref{sec:policy-checks} checks whether paired-test, task-selection, and budget-grid variants change the reporting conclusion.

Figure~\ref{fig:decision-routine} summarizes the routine used for both partial-run decisions and batch simulations. The routine first asks whether required task groups are represented. Only then does it test whether the observed paired margin is far enough from the threshold to support a final action.

\begin{figure}[t]
\scriptsize
\begin{tabular}{@{}L{0.97\columnwidth}@{}}
\textbf{Input:} visible paired outcomes \(S_b\), threshold \(\delta\), required groups, budget status, tail cutoff \(a\).\\
1. If task-group coverage fails, return \textsc{continue} if budget remains, otherwise \textsc{abstain}.\\
2. Compute \(\widehat{\Delta}_b\) and bootstrap tails \(R_{\mathrm{promote}}\), \(R_{\mathrm{reject}}\).\\
3. If budget remains, return \textsc{continue}. At the budget limit, return \textsc{promote} if \(\widehat{\Delta}_b>\delta\) and \(R_{\mathrm{promote}}\leq a\), \textsc{reject} if \(\widehat{\Delta}_b\leq\delta\) and \(R_{\mathrm{reject}}\leq a\), otherwise \textsc{abstain}.
\end{tabular}
\vspace{-1.5ex}
\caption{Decision routine used for each partial comparison.}
\label{fig:decision-routine}
\end{figure}

Once coverage holds for \(S_b\), the policy estimates uncertainty by resampling the observed paired differences \(\{D_i:t_i\in S_b\}\). Each resample draws \(|S_b|\) tasks with replacement and computes a bootstrap mean \(\widehat{\Delta}^{*}_{b}\). With bootstrap tail cutoff \(a\), define
\begin{align}
    R_{\mathrm{promote}} &= \Pr^{*}(\widehat{\Delta}^{*}_{b}\leq\delta),\\
    R_{\mathrm{reject}} &= \Pr^{*}(\widehat{\Delta}^{*}_{b}>\delta).
\end{align}
The tail cutoff is a reporting rule for the current budget cap. The policy promotes when \(\widehat{\Delta}_{b}>\delta\) and \(R_{\mathrm{promote}}\leq a\). It returns the reject action when \(\widehat{\Delta}_{b}\leq\delta\) and \(R_{\mathrm{reject}}\leq a\). Otherwise it continues if budget remains and abstains when the configured budget is exhausted. Unless otherwise stated, \(a=0.05\). The bootstrap tail is the empirical uncertainty check used inside the replay policy; repeated live looks at the same run would require a separate time-uniform sequential boundary.

A budget is sufficient only if it avoids three misleading outcomes: wrong decisions, final decisions before required task groups are represented, and a low error rate caused by leaving many comparisons unresolved. For budget \(b\), threshold \(\delta\), decision-error target \(\alpha\), coverage-failure target \(\gamma\), and abstention target \(\eta\), we require
\begin{align}
    \mathrm{FPR}_{\mathrm{cond}}(b) &\leq \alpha,\\
    \mathrm{FRR}_{\mathrm{cond}}(b) &\leq \alpha,\\
    \mathrm{coverage\ failure}(b) &\leq \gamma,\\
    \mathrm{abstain}(b) &\leq \eta.
\end{align}
Unless otherwise stated, \(\alpha=0.05\), \(\gamma=0.05\), and \(\eta=0.25\). These defaults define the paper's primary reporting policy. The 5\% decision-error targets require false promotions and false rejections to be rare, and the 5\% coverage target requires promote or reject actions before required groups appear to be rare. \(\mathrm{FPR}_{\mathrm{cond}}\) is computed over completed-record negative cases and counts false promotions; \(\mathrm{FRR}_{\mathrm{cond}}\) is computed over completed-record positive cases and counts false rejections. Abstentions remain in the relevant class denominator and are tracked separately, so these are class-conditional disagreement rates over all cases rather than selective risks among only decided cases. The 25\% abstention target prevents a rule from looking successful by deciding only easy comparisons.

Sufficient budgets are therefore reported relative to the stated policy. Different evaluation settings can choose different targets and thresholds before the run and report those choices in the decision record. We also report budgets that meet the two decision-error targets and the coverage target but still abstain too often to be useful as a decision procedure.

\section{Evaluation Protocol}

\subsection{Public Records}

ParEvalLayer needs task-level outcomes for matched tasks, not only aggregate leaderboard scores. We therefore use public records that expose outcomes for multiple agent systems, enough shared tasks to form directed pairs, and benchmark-provided groups when coverage is part of the reporting rule. Table~\ref{tab:evidence-sources} summarizes the task-count sources. Pair counts are ordered comparisons, with unordered counts in parentheses. Positive counts are ordered comparisons where the completed record promotes the tested system on the shared available overlap.

\begin{table*}[t]
    \centering
    \scriptsize
    \caption{Public task-level records used for completed-record replay of partial-evaluation decisions.}
    \label{tab:evidence-sources}
        \begin{tabular}{L{2.4cm}C{1.2cm}C{1.7cm}C{1.1cm}C{1.1cm}C{1.1cm}L{1.8cm}C{0.7cm}L{2.5cm}}
        \toprule
        Source & Pairs & Task units & Pos. 0 pp & Pos. 5 pp & Pos. 10 pp & Groups & Cost & Role in analysis \\
        \midrule
        SWE-bench Lite & 552 (276) & 300 median overlap & 275 & 199 & 154 & repository & no & software-repair task-count \\
        SWE-bench Verified & 156 (78) & 496 median overlap & 77 & 60 & 54 & repository & no & software-repair high-budget case \\
        AppWorld leaderboard & 306 (153) & 585 tasks & 153 & 128 & 108 & split by difficulty & no & app-use task-count \\
        OSWorld-Verified & 56 (28) & 359 median overlap & 28 & 27 & 21 & application & no & desktop-control task-count \\
        tau-bench public grid & 12 (6) & 278 domain-task items & 6 & 5 & 2 & domain & yes & cost-ordering case \\
        \bottomrule
    \end{tabular}
\end{table*}

The main task-count analyses use SWE-bench Lite, SWE-bench Verified, AppWorld, OSWorld-Verified, and tau-bench because they provide task outcomes and task groups suitable for the coverage rule. Terminal-Bench is evaluated separately because its public files provide wall-clock timings for usable task cells, but not the benchmark-provided task groups needed for the primary coverage rule. Section~\ref{sec:terminal-resource} asks whether stopping by elapsed time reduces unresolved comparisons enough to report. The tau-bench grid has only 12 directed pairs, so we read it as a cost-ordering check rather than a broad estimate for all conversational-agent benchmarks.

The source-specific choices are filtering, grouping, repeated-run handling, and cost availability. For SWE-bench, we keep near-complete leaderboard entries, defined as entries with outcomes for at least 95\% of the split; these records do not include per-task cost, so they support task-count budgets only. AppWorld contributes released task-level success labels and split-by-difficulty groups. OSWorld-Verified contributes application-grouped task outcomes for eight released configurations; when two runs are available for the same configuration and task, we average the task score before pairing. This treats the released record as fixed, while run-level variation is discussed in Section~\ref{sec:limits}. tau-bench contributes domain-task outcomes and cost fields, so it supports the cheap-first ordering check. We exclude banking-knowledge trajectories because retrieval configuration differs across the selected entries.

\subsection{Policies and Budget Settings}

The replay simulation fixes the reporting rule before any partial-run outcomes are computed. The rule includes the improvement threshold, source-specific task groups, task-selection policy, decision rule, budget grid, and sufficiency targets. The completed data define the comparison decision. At each budget, we reveal only the outcomes observed by that point, apply ParEvalLayer, and check promote or reject actions against the decision from the completed data.

The primary policy asks whether the observed task set covers the required task groups and is decisive enough to report. It first checks source-specific task-group coverage and then applies the bootstrap rule to decide whether the observed margin supports promote or reject. The comparison policies remove one safeguard at a time. Forced uniform and forced group-aware policies ignore uncertainty. Paired-test rules replace the bootstrap tail rule. Neyman allocation uses group variances from the completed record, and Serfling finite-population bounds use a conservative finite-population inequality. Together, these variants separate the effects of task ordering, uncertainty estimation, and prior variance information.

For SWE-bench, AppWorld, and OSWorld-Verified, we sample 500 task orders per ordered comparison and 200 bootstrap samples per policy evaluation. For tau-bench, we sample 2{,}000 task orders and 500 bootstrap samples. The main search uses budgets from 5\% to 95\% in 5 percentage point increments. We keep a small round-budget check, but the headline sufficiency claims always come from the 5 percentage point sweep.

In replay, task execution is already represented as normalized task outcomes. Partial-run and aggregate analyses use the same task rows, and completed-record fields used for after-the-fact checks remain outside the stopping rule. The stopping rule therefore sees only the outcomes available at the current task or resource budget.

\section{Results}

The results separate three cases that a partial score would conflate: partial runs whose observed outcomes already support the same conclusion as the completed evaluation, partial runs that leave too many comparisons unresolved, and checks showing that changing the budget measure or uncertainty rule does not remove the need for a decision record.

We call a budget sufficient only when the decision-error, coverage, and abstention targets in Section~\ref{sec:decision-rule} all hold; ``none by 95\%'' means no tested budget met all targets. The intervals in Table~\ref{tab:main-budget-summary} describe directed-pair variability in unresolved-comparison rates at the listed 0 pp budget; they are not additional decision criteria.

\begin{figure*}[t]
    \centering
    \includegraphics[width=0.94\textwidth]{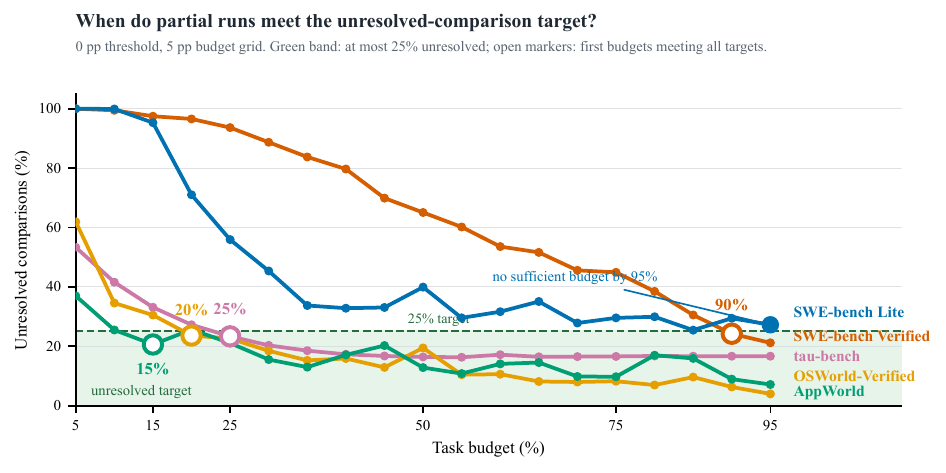}
    \caption{Unresolved comparisons at the 0 pp threshold across the 5 percentage point budget grid. Lower is better; budgets above 25\% unresolved comparisons are insufficient under the primary policy.}
    \label{fig:public-abstention-curve}
\end{figure*}

\begin{table*}[t]
    \centering
    \small
    \caption{Minimum sufficient task budgets under the primary policy.}
    \label{tab:main-budget-summary}
    \begin{tabular}{L{2.6cm}C{1.5cm}C{1.5cm}C{1.5cm}C{3.2cm}L{3.0cm}}
        \toprule
        Source & 0 pp & 5 pp & 10 pp & 0 pp unresolved interval & Result \\
        \midrule
        AppWorld & 15\% & 15\% & 15\% & 20.60\% [17.17, 23.86] & 15\% sufficient at all thresholds. \\
        OSWorld-Verified & 20\% & 20\% & 30\% & 23.61\% [16.19, 31.03] & 20\% to 30\% sufficient. \\
        tau-bench & 25\% & 30\% & 35\% & 23.35\% [7.25, 43.84] & 25\% to 35\% sufficient. \\
        SWE-bench Verified & 90\% & 90\% & 90\% & 24.22\% [20.36, 28.33] & 90\% sufficient at all thresholds. \\
        SWE-bench Lite & none by 95\% & 55\% & 35\% & 27.25\% [24.35, 30.29] & No 0 pp budget by 95\%. \\
        \bottomrule
    \end{tabular}
\end{table*}

\subsection{Replay Finds Early Decisions in Some Benchmarks}

Figure~\ref{fig:public-abstention-curve} and Table~\ref{tab:main-budget-summary} show where completed-record replay finds early decisions. At 0 pp, AppWorld reaches the reporting targets at 15\% task budget, OSWorld-Verified at 20\%, and tau-bench at 25\%. In these benchmarks, ParEvalLayer reaches the same conclusion as the completed evaluation using only 15\% to 25\% of task outcomes under the stated policy.

The early-decision pattern does not hold for every benchmark. SWE-bench Verified reaches the targets only at 90\% task budget, and SWE-bench Lite remains insufficient at 95\% under the primary repository coverage rule. The decision record makes this distinction explicit: some partial evaluations are ready to report, while others remain unresolved at the tested budget.

The limiting factor is often abstention rather than wrong decisions. SWE-bench Lite meets the decision-error and coverage targets at 25\% task budget at 0 pp, but 55.92\% of comparisons remain unresolved. Even at 95\%, 27.25\% remain unresolved, above the 25\% target. SWE-bench Verified shows the same issue at early budgets: at 25\%, the primary policy meets the error and coverage targets but leaves 93.64\% of comparisons unresolved. The abstention target is reached only at 90\%.

The threshold and task-selection policy are part of the reporting claim. SWE-bench Lite has no sufficient budget at 0 pp, but reaches 55\% at 5 pp and 35\% at 10 pp. AppWorld remains sufficient at 15\% across the three thresholds, while tau-bench and OSWorld-Verified move as the threshold changes. Because tau-bench contributes 12 directed pairs, its sufficient budgets are evidence that the policy catches a cost-ordering failure mode in this grid, not a precise benchmark-wide rate estimate. On AppWorld, the coverage-aware bootstrap policy reaches sufficiency at 15\%, but split-difficulty forced evaluation needs 50\% task budget and uniform forced evaluation fails the coverage target at 25\%. A budget number without the threshold and task-selection policy is therefore not an interpretable reporting claim.

\subsection{Partial Scores Hide Reporting Conditions}

A score-only partial report can hide whether the observed tasks are enough to stop and report the comparison. At 25\% task budget, uniform forced evaluation has 99.96\% coverage failure on both SWE-bench Lite and AppWorld, so a partial score from that subset can look precise while missing required task groups.

Task order can create a different failure. In tau-bench, cost-aware forced evaluation at 25\% task budget uses only 11.51\% of tested-system cost on average, but it has 100\% coverage failure. It also produces wrong pairwise decisions at all three thresholds, including false rejections in 4 of 5 positive 5 pp cases and 2 of 2 positive 10 pp cases. Saving cost is not enough if the observed tasks no longer support the comparison.

These examples show why unresolved cases must be reported separately from decision errors. Without that field, a partial score or an accuracy summary over only decided cases can make an incomplete comparison look ready to report.

\subsection{Resource and Decision-Rule Checks}
\label{sec:terminal-resource}
\label{sec:policy-checks}

The remaining checks ask whether the main pattern depends on measuring budget by task count or on the bootstrap uncertainty rule. Terminal-Bench lets us replace task count with elapsed time. We use public Terminal-Bench 2.0 result files \cite{merrill2026terminalbench}. After requiring at least 80 shared tasks per pair, the usable subset contains 72 directed comparisons over 89 shared terminal tasks. Wall-clock stage timing is available for all usable task cells, but native USD cost and token fields are too sparse for a cross-system cost analysis.

Elapsed-time budgets do not remove the reporting problem in this subset. At the 0 pp threshold, no budget measured by task fraction, tested-system elapsed time, or pair elapsed time meets the primary targets by 95\%. At 95\%, wrong final decisions are rare: false-promotion rates are 0.44\% to 0.62\%, and false-rejection rates are 0.00\%. The limiting factor is unresolved comparisons, which remain at 36.10\% to 36.31\%, above the 25\% target. Reports based on resource fractions still need an explicit decision rule and must state how many comparisons may remain unresolved.

The decision-rule checks point in the same direction. Relaxing the bootstrap cutoff moves some minimum budgets, as expected, but it does not turn high-budget benchmarks into early-reporting cases. On AppWorld, the sufficient budget moves from 25\% with cutoff \(a=0.02\), to 15\% with \(a=0.05\), to 10\% with \(a=0.10\). SWE-bench Verified remains a high-budget case across these cutoffs.

The main pattern is not driven by the number of bootstrap samples. When the primary coverage-aware bootstrap rule is rerun with 100, 200, and 500 bootstrap samples, the widest 5th-to-95th percentile span for unresolved comparisons across checked seeds, benchmarks, thresholds, and budgets is 1.61 pp. The largest observed conditional decision-error rate is 1.40\%, below the 5\% target. Paired-normal tests give the same sufficient-budget pattern for AppWorld, tau-bench, and SWE-bench Verified at all three thresholds; on SWE-bench Lite, they still find no sufficient 0 pp budget by 95\%.

The budget grid and comparison orientation affect how the results should be read. The 5 percentage point budget grid avoids a misleading SWE-bench Lite summary because, under the primary coverage rule, SWE-bench Lite still misses the limit on unresolved comparisons by 95\%. Directed comparisons preserve the one-sided decision question. If SWE-bench Lite is collapsed to one orientation, the 0 pp case contains 275 positive decisions and only 1 negative decision, which no longer tests both directions symmetrically.
\section{Limitations}
\label{sec:limits}

These experiments support claims about partial decisions in replayed public benchmark records, where later outcomes can be hidden after the fact. They do not by themselves establish a live stopping guarantee. A live benchmark service would also need task orders and stopping rules specified before execution, complete resource logs, explicit retry and verifier-failure policies, and enough repeated attempts to separate task variation from run variation. When public sources provide repeated runs for the same system and task, we average them before forming task pairs, so the analysis treats the released record as fixed.

The reported budgets are meaningful only through the policy recorded with the decision. The 25\% unresolved-comparison target is one such input: stricter targets would demand later budgets and looser targets would allow earlier reporting. Thresholds, task groups, coverage requirements, orientation choices, bootstrap cutoffs, and abstention targets all affect whether a partial evaluation is ready to report, so a task fraction is meaningful only with those choices. The bootstrap tail rule supports the budget comparisons in this paper; repeated live looks at the same run would require a sequential test or confidence-sequence design validated for the benchmark's sampling and dependence structure.

The uncertainty intervals should be read descriptively because ordered pairs share systems, tasks, and outcomes. Clustered or hierarchical uncertainty estimates would be stronger for future records with enough independent systems and repeated runs.

The released metadata supports coverage checks at recognizable benchmark slices. Our rule uses repository, difficulty split, application group, and domain labels to report whether a partial run skipped an important group before a comparison is reported. Finer behavioral strata or learned task clusters require pre-evaluation definitions and separate validation.

Resource claims differ by source. OSWorld-Verified supports task-count checks for the released verified configurations, Terminal-Bench supports wall-clock checks, and tau-bench supports cleaner cost fields than the other sources. Token, compute, or USD reporting decisions would require more complete resource logs across systems and tasks.

\section{Conclusion}

Partial agent evaluations should report decisions, not only scores. ParEvalLayer makes that reporting contract explicit: it reads paired task outcomes observed so far, applies a comparison policy chosen in advance, and records whether the tested agent system is better by the required amount, is not, needs more outcomes, or should abstain. Completed public records let us test whether those partial-run judgments agree with the conclusions from completed data while keeping unresolved comparisons visible.

The replay results show that the budget needed before the policy can decide varies sharply across benchmarks and policies. At the 0 pp threshold, AppWorld, OSWorld-Verified, and tau-bench can reach the same conclusion as the completed evaluation from early partial runs, while SWE-bench Verified requires much later budgets and SWE-bench Lite and the Terminal-Bench wall-clock subset remain limited by unresolved comparisons. This variation is the main empirical finding: partial evidence can be enough to report in some records, but only when the report states the decision rule and how many comparisons remain unresolved. A benchmark service can support this practice by recording the decision rule, fixing task order before execution, and keeping complete resource logs.
\bibliographystyle{IEEEtran}
\bibliography{references}

\end{document}